\documentclass[a4paper,conference]{IEEEtran}
\usepackage{times}
\usepackage{epsfig}
\usepackage{graphicx}
\usepackage{amsmath}
\usepackage{amssymb}
\usepackage{multirow}
\usepackage{dsfont}

\ifCLASSINFOpdf
\else
\fi
\begin{document}
%
\title{Temporal Binary Representation for Event-Based Action Recognition}

\author{\IEEEauthorblockN{Simone Undri Innocenti}
\IEEEauthorblockA{University of Florence \\
simone.undri@unifi.it}
\and
\IEEEauthorblockN{Federico Becattini}
\IEEEauthorblockA{University of Florence \\
federico.becattini@unifi.it}
\and
\IEEEauthorblockN{Federico Pernici}
\IEEEauthorblockA{University of Florence \\
federico.pernici@unifi.it}
\and
\IEEEauthorblockN{Alberto Del Bimbo}
\IEEEauthorblockA{University of Florence \\
alberto.delbimbo@unifi.it}}


%


\maketitle

\begin{abstract}
In this paper we present an event aggregation strategy to convert the output of an event camera into frames processable by traditional Computer Vision algorithms. The proposed method first generates sequences of intermediate binary representations, which are then losslessly transformed into a compact format by simply applying a binary-to-decimal conversion. This strategy allows us to encode temporal information directly into pixel values, which are then interpreted by deep learning models. We apply our strategy, called Temporal Binary Representation, to the task of Gesture Recognition, obtaining state of the art results on the popular DVS128 Gesture Dataset. To underline the effectiveness of the proposed method compared to existing ones, we also collect an extension of the dataset under more challenging conditions on which to perform experiments.
\end{abstract}


%
\IEEEpeerreviewmaketitle

\section{Introduction}
Action Recognition has gained increasing importance in recent years, due to applications in several fields of research such as surveillance, human computer interaction, healthcare and automotive. Despite the significant steps forward made since the diffusion of deep learning, there are still challenges yet to be solved. Certain applications, for instance, have extremely high time constraints. This is the case when recognition must be performed from fast moving vehicles (e.g. drones or cars), or when the pattern to be recognized is extremely fast (e.g. eye glimpses).
Indeed, standard RGB cameras might even fail to capture a rich enough signal to enable recognition due to low frame-rates and motion blur.

These limitations of RGB cameras have been addressed with event cameras. Event cameras, also known as neuromorphic cameras, are sensors that capture illumination changes, producing asynchronous events independently for each pixel. These sensors have several desirable properties such as high dynamic range, low latency, low power consumption, absence of motion blur and, last but not least, they operate at extremely high frequencies, generating events at a $\mu s$ temporal scale. The output of an event camera therefore is highly different from the one of a regular RGB camera, making the applicability of computer vision algorithms not so straightforward. In particular, Deep Learning methods such as Convolutional Neural Networks (CNN), work with frames of synchronous data. Asynchronous events need to be aggregated into synchronous frames to be fed to a CNN.

Several event aggregation strategies have been proposed in literature, allowing the usage of frame-based algorithms \cite{nguyen2019real, miao2019neuromorphic, ghosh2019spatiotemporal, cannici2020differentiable, cannici2019asynchronous}. These techniques however approximate the signal by quantizing time into aggregation intervals, yielding to a loss of information. The aggregation time can be lowered to limit this phenomena, but this will result in an extremely high number of frames to be processed, making real-time analysis prohibitive.

In this paper we present an event aggregation strategy named Temporal Binary Representation (TBR). Compared to existing strategies, TBR generates compact representations without losing information up to an arbitrarily small quantization time. In fact, we first aggregate events to generate intermediate binary representations with small quantization times and then losslessly combine sequences of intermediate representations into a single frame. This allows us to lower the amount of data to be processed while retaining information at fine temporal scales. TBR is specifically tailored for fast moving actions or gestures and can be directly used for training and evaluating standard CNNs. Indeed, we exploit two models based on Alexnet+LSTM and Inception 3D for action recognition, reporting state of the art results on the IBM DVS128 Gesture Dataset \cite{amir2017low}. Furthermore, we highlight the benefits of the proposed strategy by collecting an extension of the dataset in more challenging scenarios, namely higher execution speed, multiple scales, camera pose and background clutter.

To summarize, the main contributions of this paper are the following:
\begin{itemize}
    \item We propose a compact representation of event data dubbed Temporal Binary Representation, exploiting a conversion of binary event sequences into frames that encode both spatial and temporal information.
    \item Our formulation allows to tune information loss and memory footprint, making it suitable for real-time applications.
    \item We collected an extension of the popular DVS128 Gesture Dataset under challenging conditions, which we plan to release upon publication.
\end{itemize}

The paper is organized as follows: in Sec.~\ref{sec:related} a literature review is reported to frame the work in the current state of the art; in Sec.~\ref{sec:method} our Temporal Binary Representation is presented; in Sec.~\ref{sec:model} we provide an overview of the models used for classifying gestures; in Sec.~\ref{sec:dataset} we present the dataset used for evaluating our approach and introduce the additional benchmark that we have collected; in Sec.~\ref{sec:training} we discuss the training details; in Sec.~\ref{sec:experiments} and~\ref{sec:ablation} we report the results of our approach; finally in Sec.~\ref{sec:conclusions} we draw the conclusions.

\section{Related Work}
\label{sec:related}
\subsection{Action and Gesture Recognition}
Several formulations have been adopted in literature for the task of action recognition. Early works~\cite{csur2011, poppe2010} have treated it as a classification task, while more recent works have provided a finer level of detail adding a temporal dimension (action detection)~\cite{actoms,serena2016fgv,jcn2016fast,shou2017cvpr,escorcia2020guess,liu2019completeness,nguyen2019weakly} or spatial information (action localization)~\cite{ggjm2015tubes,peng2016multi,saha2016deep, singh2017online, becattini2020progress}.

Action detection aims at recognizing actions and determining their starting and ending points in untrimmed videos.
These approaches are often based on temporal proposals \cite{jcn2016fast}, i.e. a set of frame intervals that are likely to contain a generic action, which are then classified or refined \cite{shou2017cvpr, serena2016fgv}.
This concept has been extended in the spatio-temporal action localization formulation, where the temporal boundaries of the action still need to be determined, but at the same time the actor needs to be precisely localized in each frame, as in an object detection task. The output of such systems is a spatio-temporal tube \cite{ggjm2015tubes, saha2016deep, cuffaro2016segmentation}, i.e. a list of temporally adjacent bounding boxes enclosing the action.

Several works have been focusing on a specific subset of actions, referred to as gestures. Gestures can be divided into the three categories of body, hand and head gestures~\cite{mitra2007gesture}. The interest in gestures often stems from the need to establish some form of interaction between humans and machines, which indeed can happen interpreting human behaviors~\cite{liu2018gesture}. To reduce the reaction time to observed gestures, sensors with high frame-rate have been exploited~\cite{sato2006ohajiki}. Of particular interest is the usage of event cameras, which have been largely used for gesture recognition in the recent years~\cite{maro2020event, kaiser2019embodied, shrestha2018slayer, amir2017low, wang2019space, kaiser2020synaptic, ghosh2019spatiotemporal, bi2019graph}.
Some approaches rely on architectures specifically tailored to handle event data, such as spiking neural networks, which however require specialized hardware to be implemented~\cite{o2013real, kaiser2020synaptic, shrestha2018slayer}.
Most approaches, however, in order to exploit traditional computer vision algorithms, adopt an event aggregation strategy that allows the conversion of streams of asynchronous events into a set of synchronous frames. Most of these approaches, though, perform a temporal quantization in the form of histograms~\cite{ghosh2019spatiotemporal} or event subsampling~\cite{kaiser2019embodied}. To avoid information loss, the bins into which events are quantized can be shrinked, with the side effect of generating a large amount of data that has to be processed. Differently from these works, we propose an aggregation strategy that is lossless up to an arbitrarily small time interval. Our proposed approach in fact compacts several representations in a single frame, allowing to generate less data without discarding information.


\section{Event Representation}
\label{sec:method}
Events generated by an event camera are temporally and spatially localized respectively by a timestamp $t$ and pixel coordinates $x$ and $y$. Each event is also associated to a polarity $p\in \{-1, +1\}$, indicating the sign of the pixel illumination change. The output of an event camera is therefore a stream of tuples $E=(t, x, y, p)$.
To make events interpretable by standard Computer Vision algorithms, they must be aggregated into frames. In general, an aggregation algorithm is a function that maps asynchronous events into a a stream of synchronous frames.
Each generated frame $f^i$ aggregates all the events in the interval $[t^i; t^i + \Delta t ]$ spanning from an initial timestamp $t^i$ and covering a temporal extent $\Delta t$, known as accumulation time.

\subsection{Temporal Binary Representation}
Given a fixed $\Delta t$, we build an intermediate binary representation $b^i$ by simply checking the presence or absence of an event for each pixel during the accumulation time. The value in position $(x,y)$ is obtained as $b^i_{x,y} = \mathds{1}(x,y)$, where $\mathds{1}(x,y)$ is an indicator function returning 1 if an event is present in position $(x,y)$ and 0 otherwise.

\begin{figure}[!t]
	\centering
	\includegraphics[width=\columnwidth]{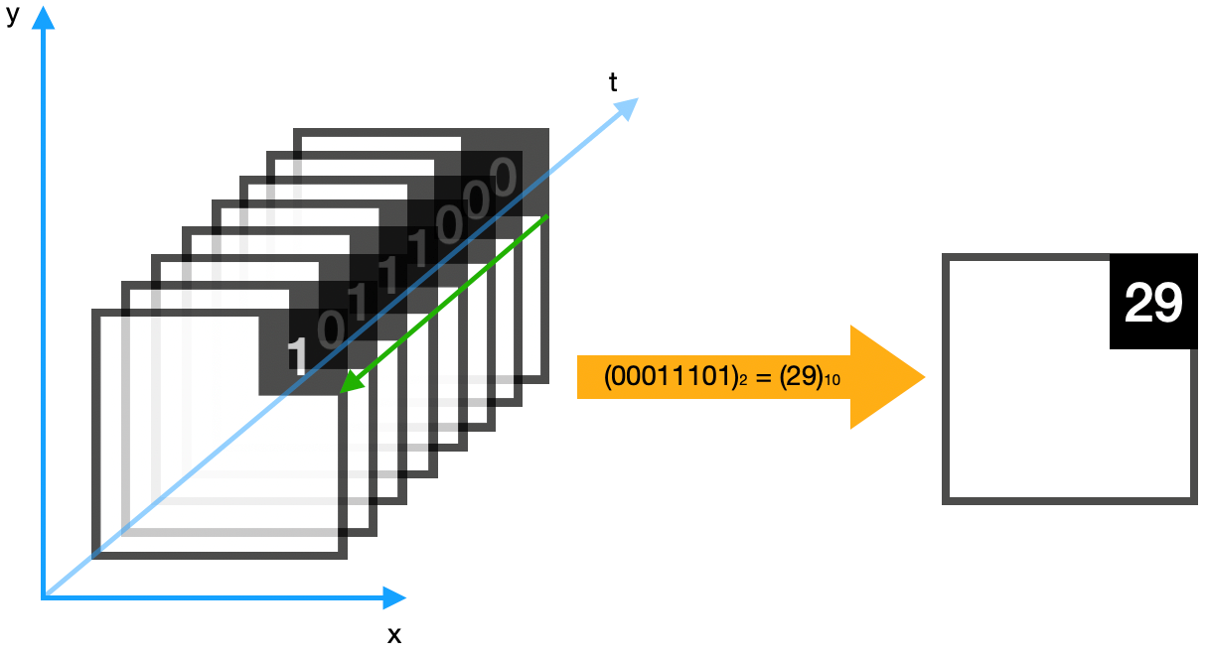}
	\caption{Temporal Binary Representation. Events are first stacked together into intermediate binary representations which are then grouped into a single frame thanks to a binary to decimal conversion.}
	\label{fig:binaryevent}
\end{figure}

We then consider $N$ temporally consecutive binary representations by stacking them together into a tensor $B \in \mathbb{R}^{H \times W \times N}$. Each pixel can be considered as a binary string of $N$ digits $[b^0_{x,y}~ b^1_{x,y}~ ...~ b^{N-1}_{x,y}]$ with the most significant digit corresponding to the most recent event. We then convert into a decimal number each binary string, as shown in Fig. \ref{fig:binaryevent}. This procedure allows us to compact the representation of $N$ consecutive accumulation times into a single frame without any loss of information. The frame is then normalized in $[0, 1]$, dividing its values by $N$.
We refer to this event representation as Temporal Binary Representation (TBR).

Compared to standard event aggregation strategies that generate a single frame for each $\Delta t$, TBR reduces the memory footprint by a factor of $N$. This also leads to less data to be processed by Computer Vision algorithms, enabling time-constrained applications. At the same time, the accumulation time can be significantly reduced to capture events at finer temporal scales, without increasing the total number of frames.

\begin{figure*}[t]
	\centering
	\includegraphics[width=0.9\columnwidth]{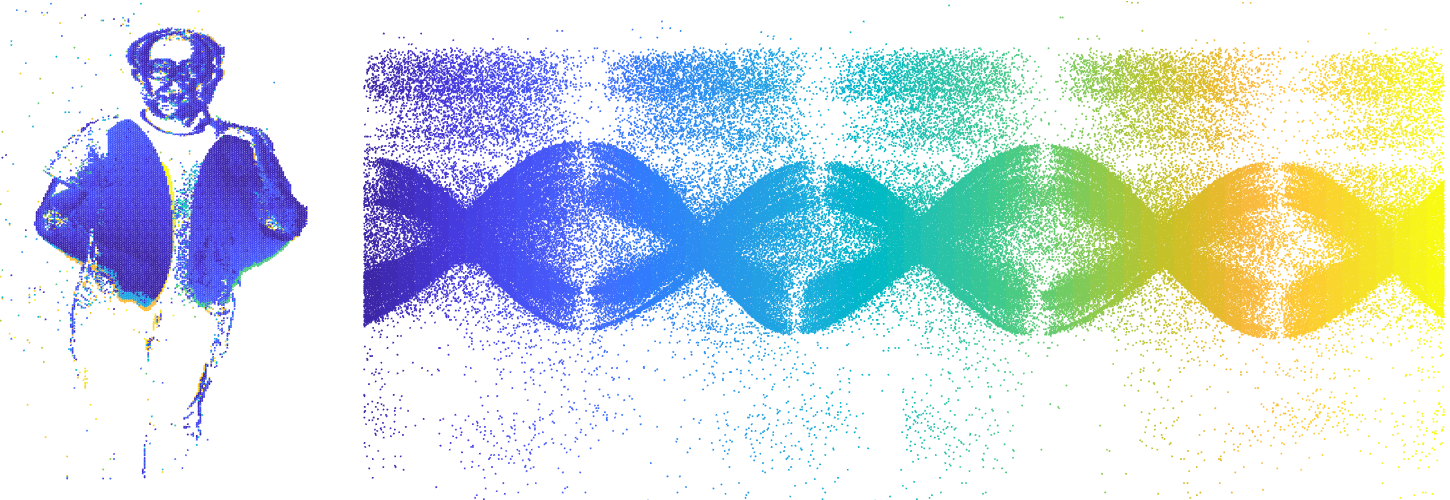}
	\includegraphics[width=0.9\columnwidth]{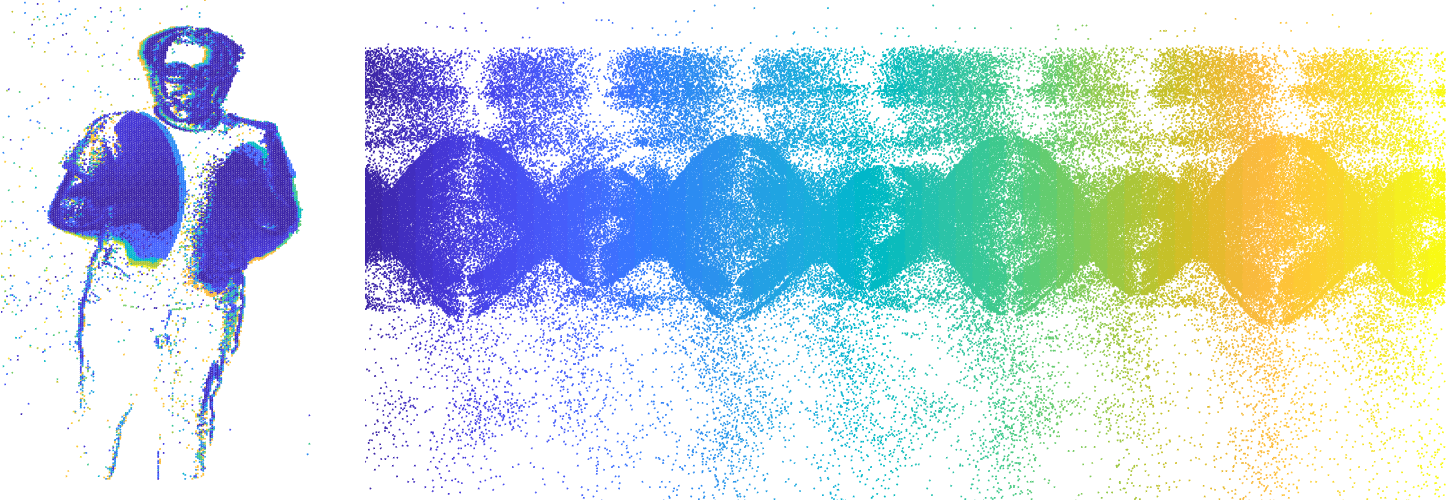}\\
	\includegraphics[width=0.9\columnwidth]{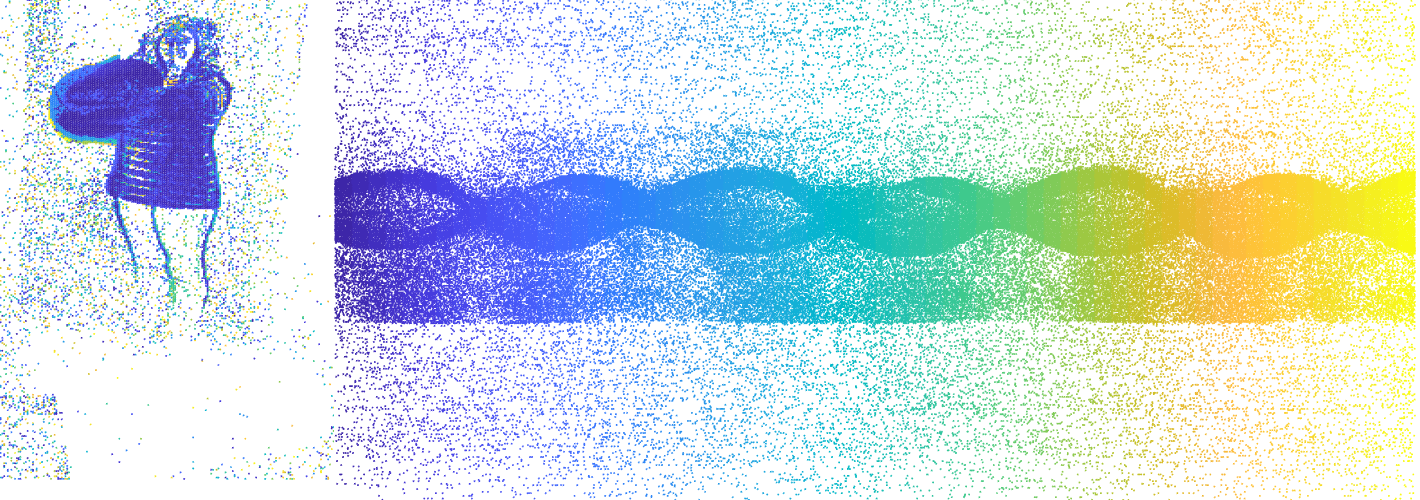}
	\includegraphics[width=0.9\columnwidth]{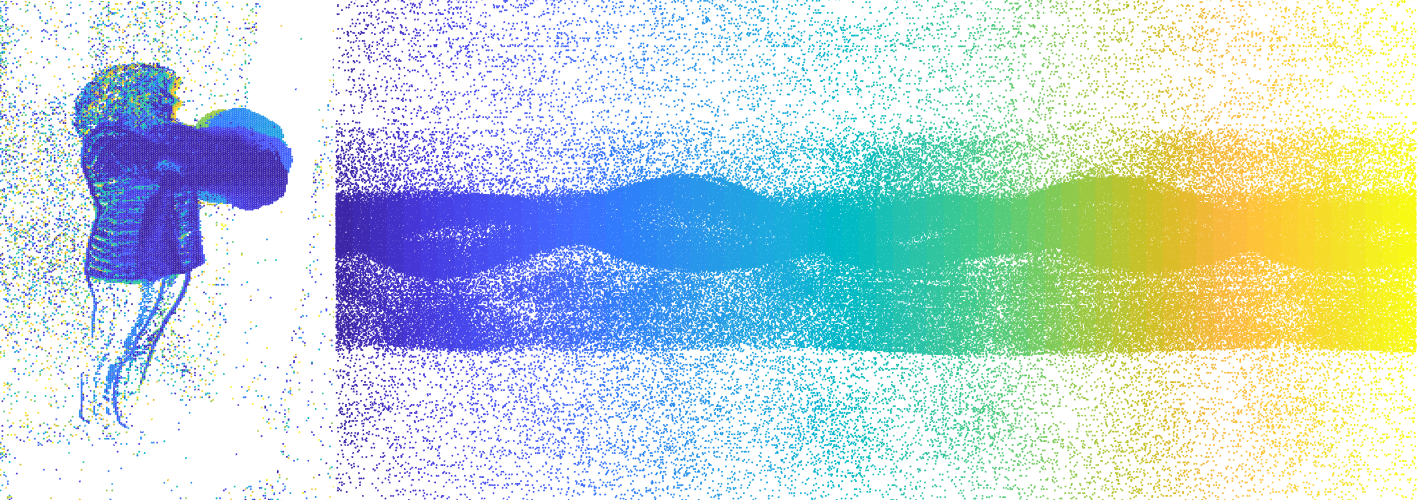}\\
	\caption{Samples from the MICC-Event Gesture Dataset. Slow and fast execution of the action \textit{air drum} (first row) and different scale and orientation of the action \textit{arm roll} (second row). A 1 second snippet is shown for each sample, where events are color-coded according to the timestamps from blue (start - 0s) to yellow (end - 1s). The actors are shown both frontal (left) and sideways (right).}
	\label{fig:recordings}
\end{figure*}

\section{Model}
\label{sec:model}
We adopt our Temporal Binary Representation for event camera data to the task of Action Recognition. To process frames, we use two different architectures.

First, we combine a Convolutional Neural Network to extract spatial features with a Recurrent Neural Network to process sequences of frames. As a feature extractor, we train an AlexNet~\cite{krizhevsky2012imagenet}, replacing the final fully connected layers using a single layer with 512 neurons. The output for each frame in the sequence is directly fed to a Long Short Term Memory (LSTM) with 2 layers with hidden dimension 256 each. Finally, a fully connected layer with softmax activation performs the classification.

The second model that we adopt is the Inception 3D model \cite{carreira2017quo}, a state of the art architecture widely adopted with RGB data for action recognition. Based on Inception-V1 \cite{szegedy2015going}, the model relies on inflated 3D convolutions by adding a third dimension to filters and pooling kernels to learn spatio-temporal feature extractors. The model originally has two separate streams for RGB and Optical Flow data. Here we simply remove one branch and retrain the model with event camera data aggregated with TBR.

To process videos, we follow two different approaches, depending on the network. For the AlexNet+LSTM model we simply feed the whole sequence of frames to the model and collect the final output. With Inception 3D instead, we use as input non-overlapping blocks of $F$ frames stacked together, which are independently evaluated. To provide a final classification for the whole video, we adopt a majority-voting strategy among predictions for each block.


\section{Datasets}
\label{sec:dataset}
We train our model on the the IBM DVS128 Gesture Dataset~\cite{amir2017low}. The dataset contains a total of 1342 hand gestures with a variable duration spanning from approximately 2 to 18 seconds (6 seconds on average). Gestures are divided in 10 classes plus an additional random class for unknown gestures. Each of these actions are performed by 29 subjects under different illumination conditions (natural, fluorescent and led lights). The data is acquired using a DVS128 camera, i.e. an event camera with a sensor size of $128 \times 128$ pixels~\cite{lichtsteiner2006128}. We follow the split proposed by the authors, comprising 23 subjects for training and 6 for validation.

To increase the variability of the DVS128 Gesture Dataset we recorded an additional test benchmark using a Prophesee GEN 3S VGA-CD event camera\footnote{https://www.prophesee.ai/event-based-evk/}. The camera has a sensor with a higher resolution of $640 \times 480$ pixels (VGA). The recorded actions still belong to the 11 classes of the DVS128 dataset but are performed under more challenging conditions. In particular, the actors were asked to perform the actions at different speeds, in order to demonstrate the capacity of event cameras to capture high speed movements. In addition the actions have been recorded at different scales and camera orientations and also under uneven illumination which is likely to cast shadows on the body and the surroundings, generating spurious events.
The dataset was recorded by 7 different actors of different age, height and gender for a total of 231 videos. All the videos are used for testing, still using the DVS128 Gesture Dataset as training set.
We refer to the newly collected data as the MICC-Event Gesture Dataset, which will be released upon publication.
In Fig.~\ref{fig:recordings} a few samples from the dataset are shown, highlighting the different execution speeds, scales and orientations at which actions are recorded.

\section{Training}
\label{sec:training}
We train the models using the SGD optimizer with momentum. We use a learning rate equal to 0.01, which is then decreased to 0.001 after 25 epochs. As loss we adopt the Binary Cross-Entropy Loss, regularized with weight decay. Overall, the training of Inception 3D took 13 hours on an NVIDIA Titan Xp, while AlexNet+LSTM required approximately 30 hours.

For the DVS128 Gesture Dataset, to make the frames compatible with the input layer of the models, we apply a zero-padding up to $227\times227$ for AlexNet+LSTM and $224 \times224$ for Inception 3D. For the MICC-Event Gesture Dataset instead, which is recorded with the higher resolution of $640 \times 480$, we perform a central crop of $350 \times 350$ pixels and then reshape it to $128 \times 128$ to match the size of DVS128. Reshape is done with Nearest Neighbor interpolation to a avoid unwanted artifacts that may introduce noise in the event representation.
Frame values are normalized in $[-1; 1]$ before being fed to the models.
During training we also perform data augmentation applying random scaling, translation and rotation.

\section{Experiments}
\label{sec:experiments}
In Tab.~\ref{tab:dvs128} we report the results on the DVS128 Gesture Dataset for the two models AlexNet+LSTM and Inception 3D, trained with frames generated by our Temporal Binary Representation. The results are compared with state of the art approaches. Following prior work, we report the classification accuracy both including and excluding the \textit{Other Gesture} class, respectively referred to as "10 classes" and "11 classes".

In our models, events are aggregated with the proposed Temporal Binary Encoding, stacking $N=8$ binary representations with an accumulation time $\Delta t=2.5 ms$. Therefore, we use an 8 bit representation for each pixel, covering 20 ms with each frame. It is important to notice that this allows the model to observe events without any loss of information up to a precision of 2.5 ms, even if a single frame stores data covering an 8 times bigger time interval.
Since the Inception 3D model takes as input chunks of videos as a tensor of temporally stacked frames, we feed to the model chunks of 500 ms, i.e. chunks of 25 frames encoded with TBR. With classic event aggregation strategies that use the same $\Delta t$ of 2.5 ms, this would lead to 200 frames per chunk, increasing considerably the computational burden.

Overall, the Inception 3D model achieves the best results, reporting approximately a 2\% improvement respect to AlexNet+LSTM. Interestingly, both our architectures are capable to obtain a perfect classification of the \textit{Other Gesture} class, making the accuracy in the 11 classes settings higher than the 11 classes one. This behavior is the opposite compared to the baselines that adopt the 10 classes setting, which consistently lowers the accuracy.

\begin{table}[]
\caption{Results on the DVS128 Gesture Dataset.}
\label{tab:dvs128}
\centering
\begin{tabular}{l|c|c}
                                                    & 10 classes                & 11 classes                     \\ \hline
Time-surfaces~\cite{maro2020event}                  & 96.59                     & 90.62                          \\
SNN eRBP\cite{kaiser2019embodied}                   & -                         & 92.70                          \\
Slayer~\cite{shrestha2018slayer}                    & -                         & 93.64                          \\
CNN~\cite{amir2017low}                              & 96.49                     & 94.59                          \\
Space-time clouds~\cite{wang2019space}              & 97.08                     & 95.32                          \\
DECOLLE~\cite{kaiser2020synaptic}                   & -                         & 95.54                          \\
Spatiotemporal filt.~\cite{ghosh2019spatiotemporal} & -                         & 97.75                          \\
RG-CNN~\cite{bi2019graph}                           & -                         & 97.20                          \\ \hline
Ours - AlexNet+LSTM                                 & 97.50                     & 97.73                          \\
Ours - Inception3D                                  & \textbf{99.58}                     & \textbf{99.62}                 
\end{tabular}
\end{table}


To better assess the benefits of adopting our Temporal Binary Representation, we report results on the MICC-Event Gesture Dataset. We use the whole dataset for testing the Inception 3D model, which is trained on DVS128. To provide a comparison with other approaches, we have trained 2 baseline variants using event aggregation strategies from the literature: \textit{Polarity}~\cite{nguyen2019real} and \textit{Surface of Active Events}~\cite{mueggler2017fast}.

The \textit{Polarity}~\cite{nguyen2019real} approach simply assigns a different value to events with different polarities. Therefore, the final representation is an image $I_p$, where each pixel $(x,y)$ is given by:
\begin{equation}
I_{p}(x,y) = 
\begin{cases}
0, \hspace{15px}  \text{if event polarity is negative}\\
0.5, \hspace{7px} \text{if no events happen in $\Delta$t}\\
1, \hspace{15px} \text{if event polarity is positive}
\end{cases}
\end{equation}
If multiple events are detected in the accumulation time, the most recent one is considered.

The \textit{Surface of Active Events} (SAE)~\cite{mueggler2017fast} instead, for each pixel measures the time between the last observed event and the beginning of the accumulation time $t_0$. The values are then normalized between 0 and 255, similarly to TBR with 8 bits. Polarity is discarded. The representation $I_{SAE}$ is obtained as:
\begin{equation}
I_{SAE}(x, y) = 255 \times \left(\frac{\text{t\scriptsize{p}} - \text{t\scriptsize{0}}}{\text{$\Delta$t}}\right)
\end{equation}

\begin{figure}[!t]
	\centering
	\includegraphics[width=0.3\columnwidth]{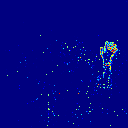}
	\includegraphics[width=0.3\columnwidth]{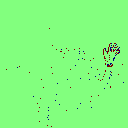}
	\includegraphics[width=0.3\columnwidth]{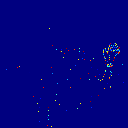} \\ \medskip
	
	\includegraphics[width=0.3\columnwidth]{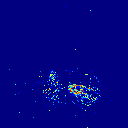}
	\includegraphics[width=0.3\columnwidth]{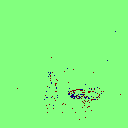}
	\includegraphics[width=0.3\columnwidth]{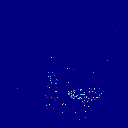}
	
	\caption{Events aggregated with our Temporal Binary Representation (left), Polarity~\cite{nguyen2019real} (middle) and Surface of Active Events~\cite{mueggler2017fast} (right). All three representations are made using an accumulation time $\Delta t=2.5 ms$.}
	\label{fig:accumulation_strategies}
\end{figure}

Samples using TBR, Polarity and SAE are shown in Fig.~\ref{fig:accumulation_strategies}.

\begin{table}[t]
	\caption{Results on the DVS128 Gesture Dataset and the MICC-Event Gesture Dataset for Inception 3D trained with three different aggregation strategies: TBR (ours), Polarity~\cite{nguyen2019real} and SAE~\cite{mueggler2017fast}.}
	\label{tab:micc-event}
	\begin{tabular}{l|c|c|c}
		& TBR (ours) & Polarity & SAE   \\ \hline
		DVS128 Gesture Dataset     & \textbf{99.62}		&	98.86		& 98.11 \\
		MICC-Event Gesture Dataset & \textbf{73.16}		& 	68.40		& 70.13 \\
	\end{tabular}
\end{table}

\begin{figure*}[t]
	\centering
	\begin{tabular}{ccccc}
		$\Delta t=1 ms$ & $\Delta t=2.5 ms$ & $\Delta t=5 ms$ & $\Delta t=10 ms$ & $\Delta t=20 ms$ \\
		\includegraphics[width=0.15\textwidth]{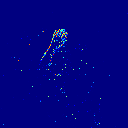} &
		\includegraphics[width=0.15\textwidth]{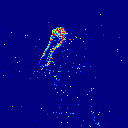} &
		\includegraphics[width=0.15\textwidth]{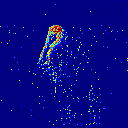} &
		\includegraphics[width=0.15\textwidth]{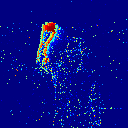} &
		\includegraphics[width=0.15\textwidth]{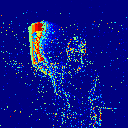} \\
		
		\includegraphics[width=0.15\textwidth]{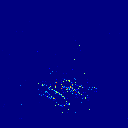} &
		\includegraphics[width=0.15\textwidth]{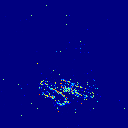} &
		\includegraphics[width=0.15\textwidth]{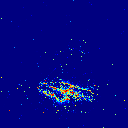} &
		\includegraphics[width=0.15\textwidth]{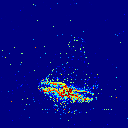} &
		\includegraphics[width=0.15\textwidth]{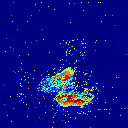} \\
		
		\includegraphics[width=0.15\textwidth]{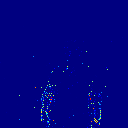} &
		\includegraphics[width=0.15\textwidth]{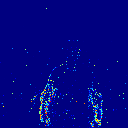} &
		\includegraphics[width=0.15\textwidth]{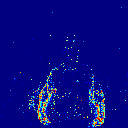} &
		\includegraphics[width=0.15\textwidth]{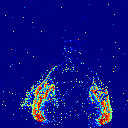} &
		\includegraphics[width=0.15\textwidth]{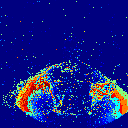} \\
	\end{tabular}
	\caption{Temporal Binary Representations with different accumulation times $\Delta t$ with a number of bits $n=8$. Each frame represents all the events in the interval $[ 0; n \times \Delta t ]$. Three different gestures are shown: \textit{Right Hand Clockwise} (top); \textit{Arm Roll} (middle); \textit{Other Gesture} (bottom). Pixels are color-coded according to intensity, from 0 (blue - no events) to 255 (red - an event registered for each bit of the representation).}
	\label{fig:deltat_fig}
\end{figure*}

In Tab.~\ref{tab:micc-event} we show the results obtained by Inception 3D trained with the three different aggregation strategies. All three strategies are used with an accumulation time $\Delta t=2.5 ms$. We also report the results on the original DVS128 Gesture Dataset test set obtained by our model with the baseline aggregation strategies. Interestingly, on DVS218 the three variants still obtain higher performances than the existing methods from the literature reported in Tab.~\ref{tab:dvs128}. This confirms the choice of Inception 3D, which proves to be suitable for the task of action/gesture recognition using event data.

The results on the MICC-Event Gesture Dataset overall are much lower due to the challenging scenarios that we have collected. However, the gap between the proposed aggregation strategy and the baselines increases considerably, suggesting that the Temporal Binary Representation is capable of representing event data more effectively. At the same time, since we are using $N=8$ bits, TBR generates 8 times less data to be processed since N frames are losslessly condensed into a single representation.

\section{Ablation Studies}
\label{sec:ablation}
We perform a series of ablation studies, showing the performance of the proposed method varying the parameters of the Temporal Binary Representation strategy.
In particular, we observe how the accuracy of the system is affected when varying the accumulation time $\Delta t$, the number of bits used for the binary representation and the length of the video chunk fed to the Inception 3D model.

\subsection{Accumulation time}
\begin{figure}[!t]
	\centering
	\includegraphics[width=0.9\columnwidth]{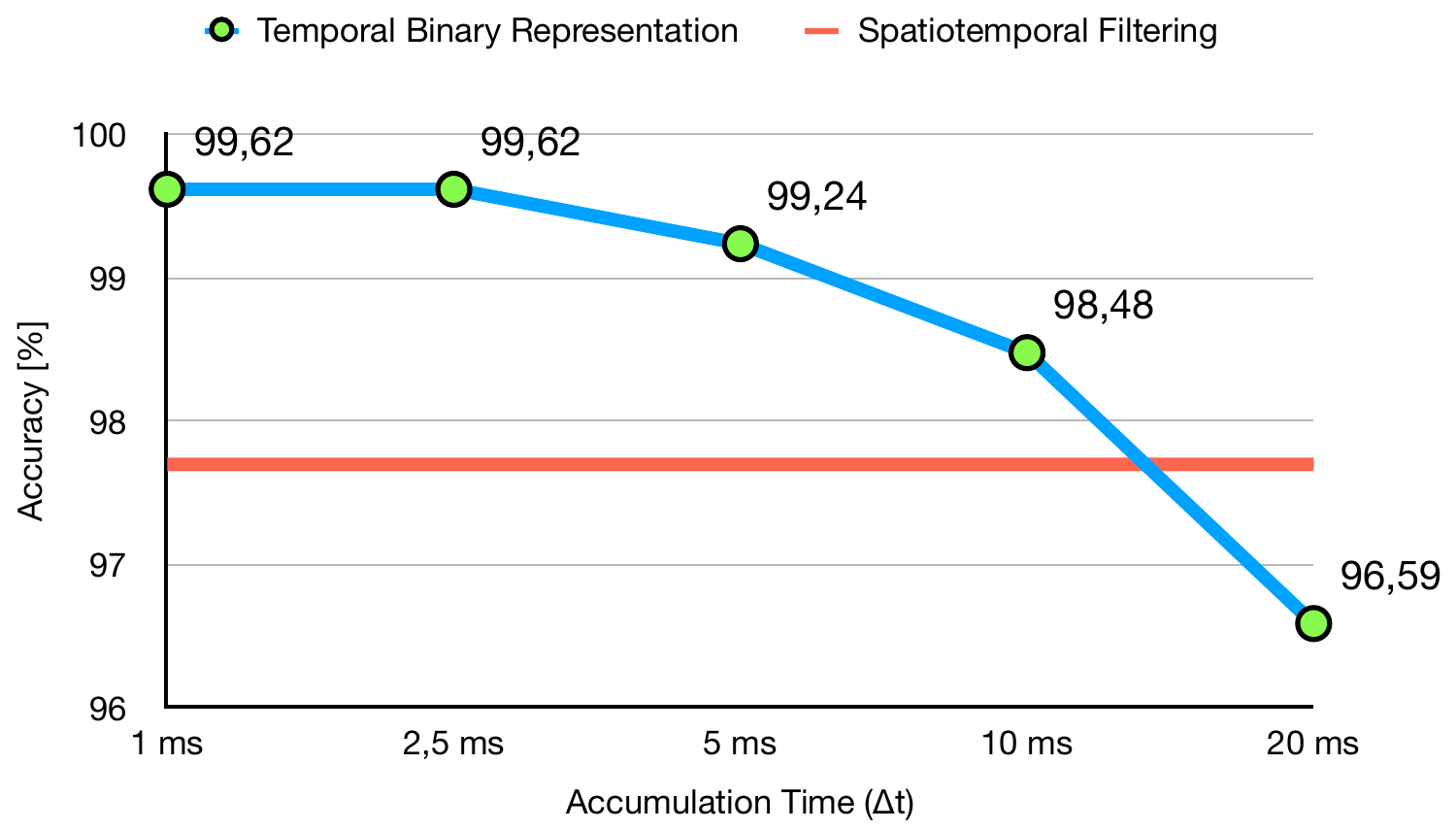}
	\caption{Accuracy of Inception 3D on the DVS 128 Gesture Dataset, varying the accumulation time $\Delta t$. The best results from the state of the art~\cite{ghosh2019spatiotemporal} is also shown as reference.}
	\label{fig:deltat}
\end{figure}

Varying the accumulation time $\Delta t$, we can adjust the temporal quantization made by TBR. Higher accumulation times lead to more compact representations, which however carry less information. It can be seen from Fig.~\ref{fig:deltat} that this information loss comes with a drop in accuracy for accumulation times bigger than 2.5 ms. Interestingly, lowering $\Delta t$ beneath this threshold does not bring any improvement for the task at hand. In the plot, the best result from the state of the art~\cite{ghosh2019spatiotemporal}, is shown as reference.

Fig.~\ref{fig:deltat_fig} shows samples of Temporal Binary Representations for different accumulation times. Especially for sufficiently high $\Delta t$, both the spatial and temporal nature of the encoding appears clearly visible.

\subsection{Number of bits}
\begin{figure}[!t]
	\centering
	\includegraphics[width=0.9\columnwidth]{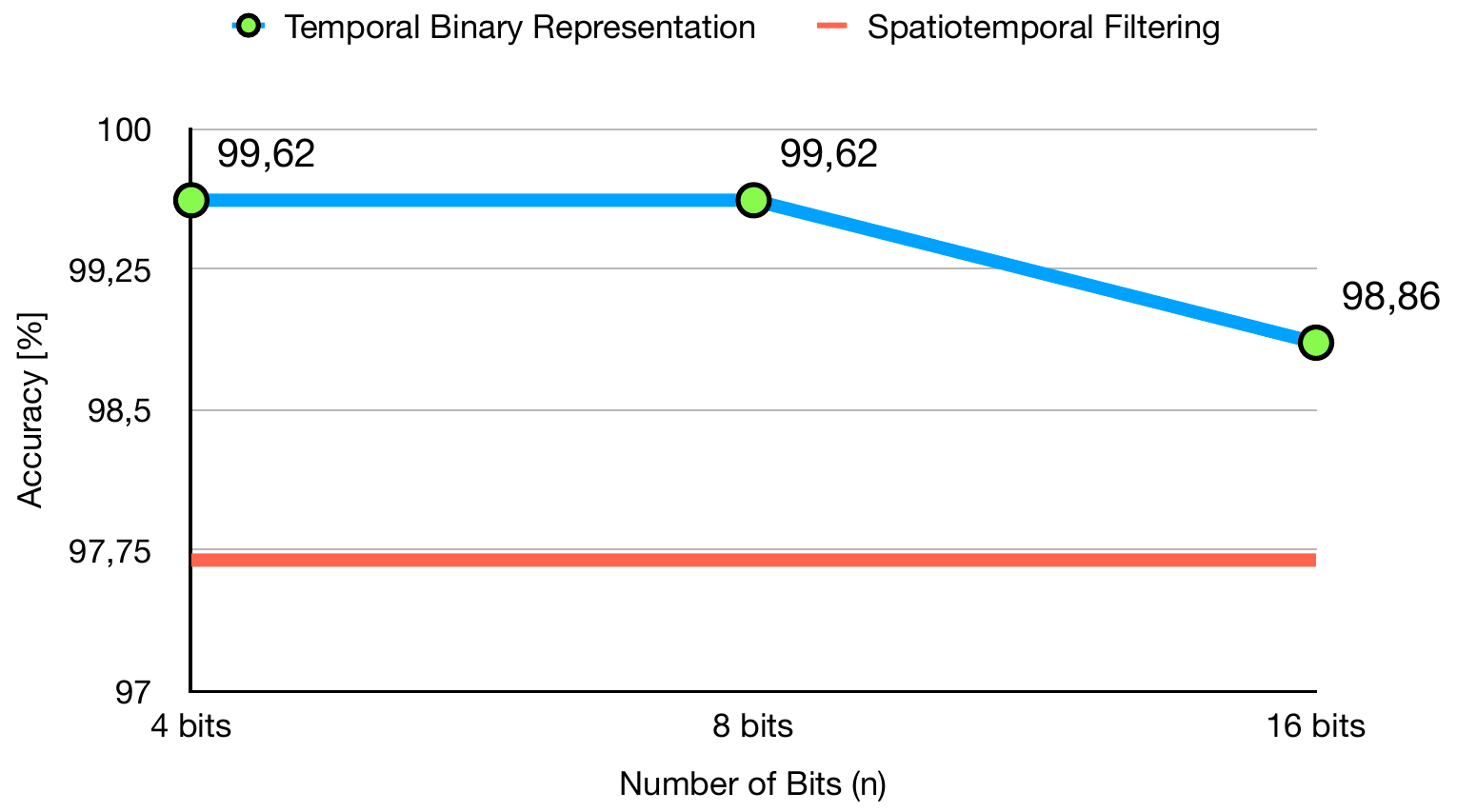}
	\caption{Accuracy of Inception 3D on the DVS 128 Gesture Dataset, varying the number of bits for the Temporal Binary Representation. The best results from the state of the art~\cite{ghosh2019spatiotemporal} is also shown as reference.}
	\label{fig:n}
\end{figure}
Along with $\Delta t$, the number of bits $N$ used for the proposed Temporal Binary Representation, defines how much information gets condensed into a single frame. Fig.~\ref{fig:n} shows the accuracy of Inception 3D on the DVS128 Gesture Dataset using $N={4,8,16}$. Similarly to $\Delta t$, when $N$ becomes too small, the accuracy of the model saturates. Throughout the paper we have taken $N=8$ bits as reference for building our representations since it offers a trade-off between accuracy and data compactness. Furthermore, the choice of $N=8$ simplifies data storage since events can be saved as unsigned integers grayscale images with lossless compression.

\subsection{Chunk length}
\begin{figure}[t]
	\centering
	\includegraphics[width=0.9\columnwidth]{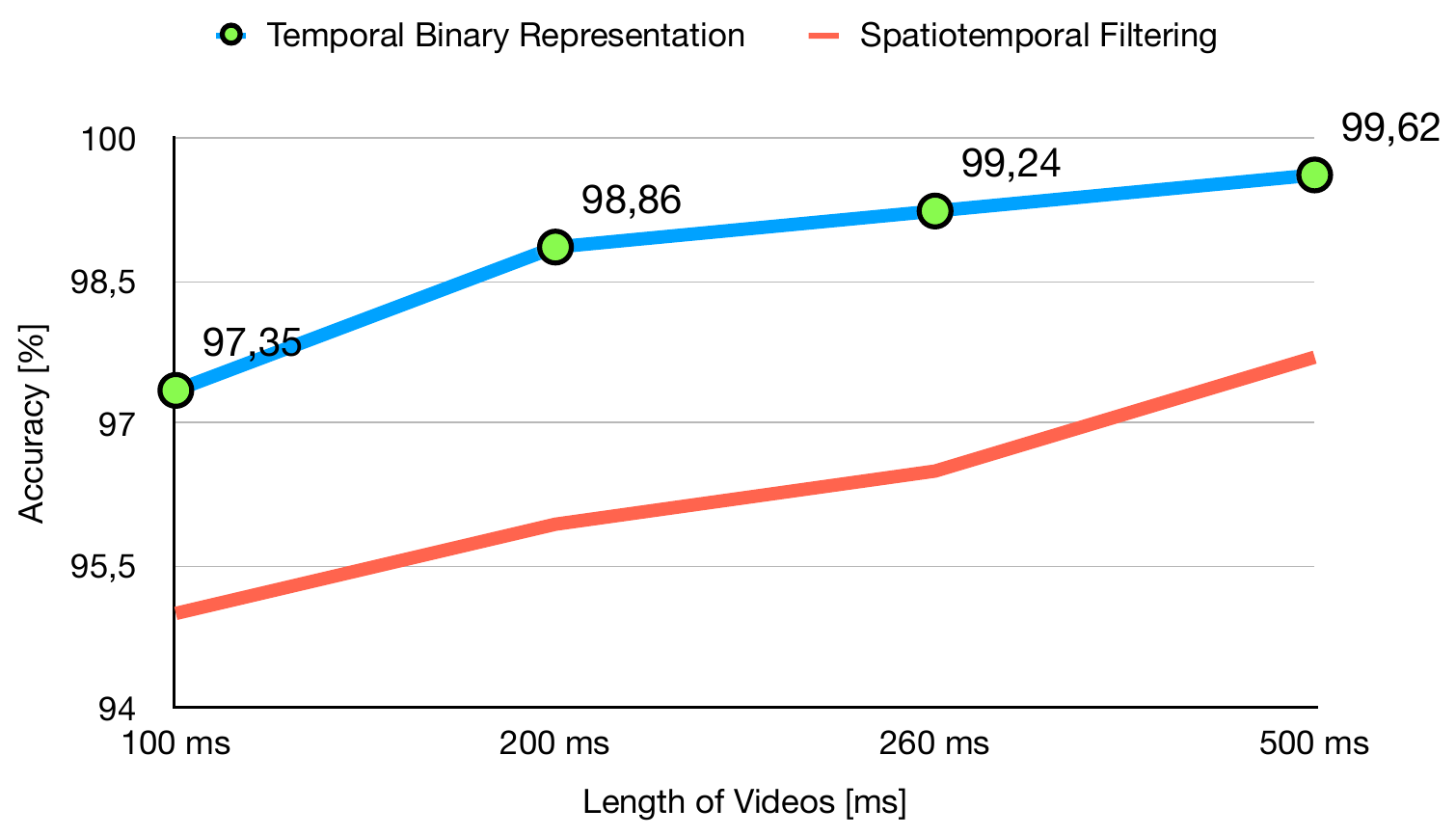}
	\caption{Accuracy of Inception 3D on the DVS 128 Gesture Dataset, varying the chunk size fed to Inception 3D. The best results from the state of the art~\cite{ghosh2019spatiotemporal} is also shown as reference.}
	\label{fig:chunk}
\end{figure}
Here we vary the length of the chunks fed to Inception 3D. Since the model exploits 3D inflated convolutions, it can process multiple frames concatenated together, therefore taking into account the temporal dimension. In the case of TBR, the temporal dimension is already encoded covering a timespan of $N\times\Delta t$.
By staking frames together we are extending the observation timespan by a factor equal to the number of frames. This setting is equivalent to the one adopted in \cite{ghosh2019spatiotemporal}, where the classifier performs a majority voting after having observed several chunks of various dimensions.
In Fig.~\ref{fig:chunk}, we report the results for both methods, varying the chunk length from 100 ms to 500 ms. For our Temporal Binary Encoding we use $\Delta t=2.5 ms$ and $N=8$, hence covering with each frame a temporal interval of 20 ms.
The accuracy of the system improves when the chunk length increases, up to 500 ms. We did not observe significant improvements when increasing it further by adding more frames. It has to be noted however that increasing the chunk length will also increase the latency of the model, since a longer part of the gesture needs to be observed before emitting the first classification.

\section{Conclusions}
\label{sec:conclusions}
In this paper we have presented an accumulation strategy called Temporal Binary Representation for converting the output of event cameras from raw events to frames, making them processable by Computer Vision algorithms. The proposed approach generates highly compact data, thanks to a lossless conversion of intermediate binary representations into a single decimal one. The effectiveness of the proposed approach has been validated on the commonly used DVS128 Gesture Dataset, reporting state of the art results. In addition a new test benchmark for event-based gesture recognition has been collected and will be publicly released.

\section*{Acknowledgments}
This work was partially supported by the Italian MIUR within PRIN 2017, Project Grant 20172BH297: I-MALL - improving the customer experience in stores by intelligent computer vision.



%
%
%

\bibliographystyle{IEEEtran}
\bibliography{egbib}

\end{document}